\documentclass[letterpaper, 10 pt, conference]{ieeeconf} 

\IEEEoverridecommandlockouts      
\usepackage{times}

\usepackage{multicol}   

\usepackage{amsmath} 
\usepackage{amssymb}  

\usepackage{amsthm}
\usepackage{dsfont}

\usepackage{bm}
\usepackage{booktabs}
\usepackage{subcaption}

\usepackage{cleveref} 
\usepackage{dsfont}
\usepackage{graphicx}
\usepackage{psfrag,graphicx,epsfig}

\usepackage{epstopdf}
\usepackage{xspace}
\usepackage{subcaption}
\usepackage{float}
\usepackage{placeins}
\usepackage{multirow}
\usepackage{pgf,tikz}
\usepackage{nowidow}
\usepackage{lineno}
\usepackage{xcolor}
\usepackage{color}
\usepackage{siunitx}
\usepackage{color}
\usepackage{lineno}
\usepackage{algorithmic}
\usepackage{gensymb}
\usepackage{adjustbox}
\usepackage{cuted}

\usepackage{siunitx}
\usepackage{color}
\usepackage{flushend}
\usepackage{lineno}
\usepackage{gensymb}
\usepackage{algorithm,verbatim}
\usepackage{color-edits}
\addauthor{ms}{magenta}
\addauthor{jw}{green}
\addauthor{tp}{red}
\addauthor{tz}{orange}
\addauthor{xn}{purple}
\addauthor{hz}{brown}

\theoremstyle{definition}

\theoremstyle{remark}

\allowdisplaybreaks

\begin{document}

\title{
\LARGE \bf
Analytic Conditions for Differentiable Collision Detection \\in Trajectory Optimization
}

\author{Akshay Jaitly$^{1}$, Devesh K. Jha$^{2}$, Kei Ota$^{3}$,  and Yuki Shirai$^{2}$ %
\thanks{$^{1}$A. Jaitly is with the Robotics Engineering Department, Worcester
Polytechnic Institute, Worcester, MA 01609, USA {\tt\small akshayjaitly@hotmail.com}}
\thanks{$^{2}$D. K. Jha and Y. Shirai are with Mitsubishi Electric Research Laboratories, Cambridge, MA, USA 02139 {\tt\small \{jha,shirai\}@merl.com}}
\thanks{$^{3}$K. Ota is with Mitsubishi Electric, Kanagawa, Japan {\tt\small Ota.Kei@ds.MitsubishiElectric.co.jp}}
}

\maketitle




\begin{abstract}
Optimization-based methods are widely used for computing fast, diverse solutions for complex tasks such as collision-free movement or planning in the presence of contacts. However, most of these methods require enforcing non-penetration constraints between objects, resulting in a non-trivial and computationally expensive problem. This makes the use of optimization-based methods for planning and control challenging. In this paper, we present a method to efficiently enforce non-penetration of sets while performing optimization over their configuration, which is directly applicable to problems like collision-aware trajectory optimization. We introduce novel differentiable conditions with analytic expressions to achieve this. To enforce non-collision between non-smooth bodies using these conditions, we introduce a method to approximate polytopes as smooth semi-algebraic sets. We present several numerical experiments to demonstrate the performance of the proposed method and compare the performance with other baseline methods recently proposed in the literature.


\end{abstract}

\section{Introduction}

Optimization-based approaches present an effective way to generate rich behavior for robots in the presence of various kinds of constraints~\cite{bostondynamicsPickingMomentum, kelly2017introduction, chandler2024galileo}. These methods are widely used for planning trajectories of multi-body robotic systems in the presence of obstacles, or in the presence of contact constraints \cite{manchester2020variational}. They are also used in various physics engines 
(e.g., MuJoCo \cite{todorov2012mujoco}, Bullet \cite{coumans2015bullet}, Drake \cite{drake}, Dojo \cite {howell2022dojo}, 
PhysX \cite{nvidiaPhysX})
where simulation of contact dynamics is performed by first finding the contacts between objects, then solving a constrained optimization problem. 

Central to these planning or simulation problems is the ability to compute a signed distance function between bodies which can later be used for downstream tasks like collision-free trajectory planning or simulating contact dynamics. Computation of distance functions tends to be computationally challenging. Oftentimes, evaluating constraints on collision or distance values is non-differentiable, leading to challenges when solving problems in various applications like computer graphics, robotics, video games, etc. 


\begin{figure}[t] 
  \centering
\begin{subfigure}{\columnwidth}
\includegraphics[width=\columnwidth]{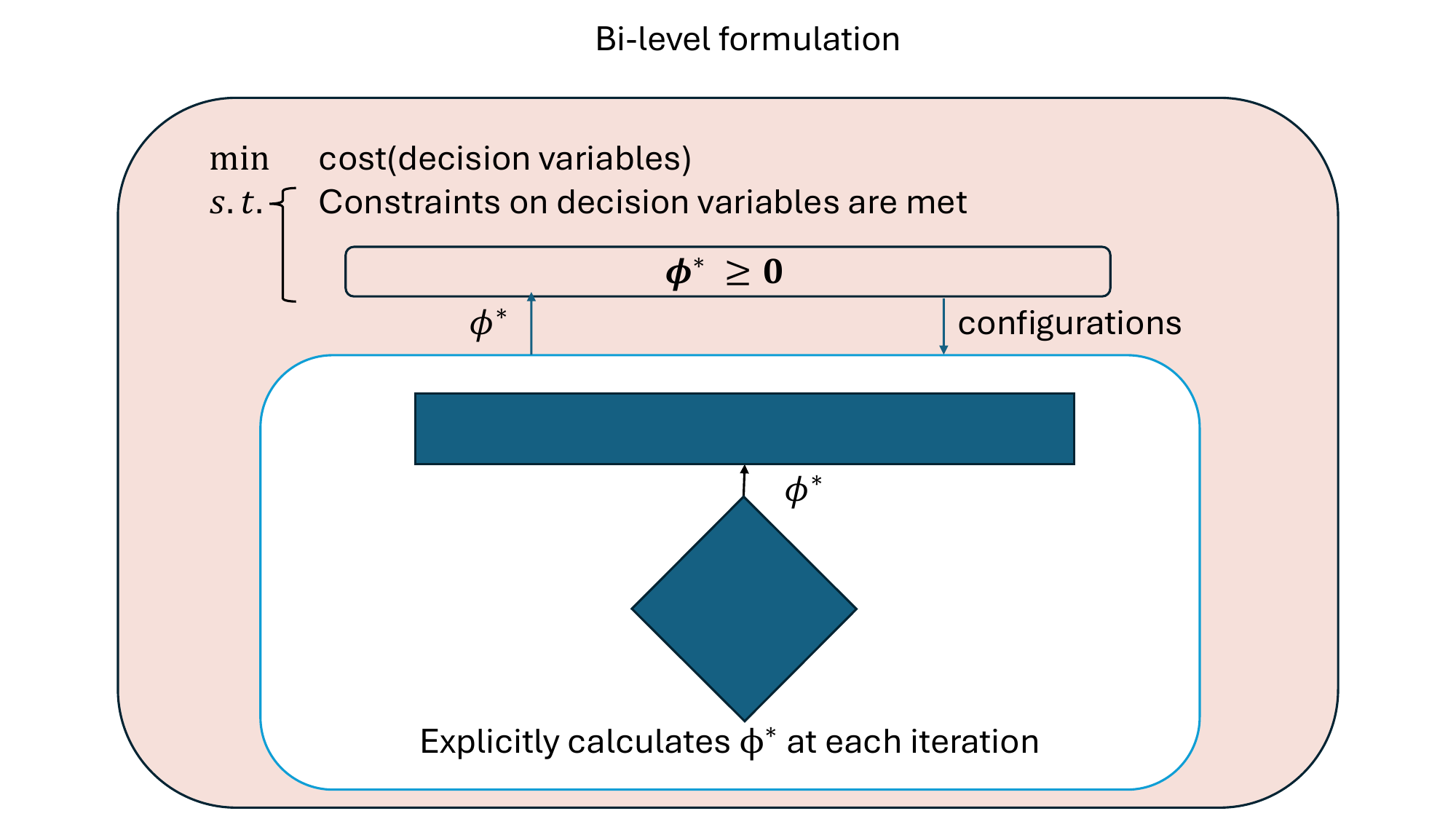}
\end{subfigure}\quad
\begin{subfigure}{\columnwidth}
\includegraphics[width=\columnwidth]{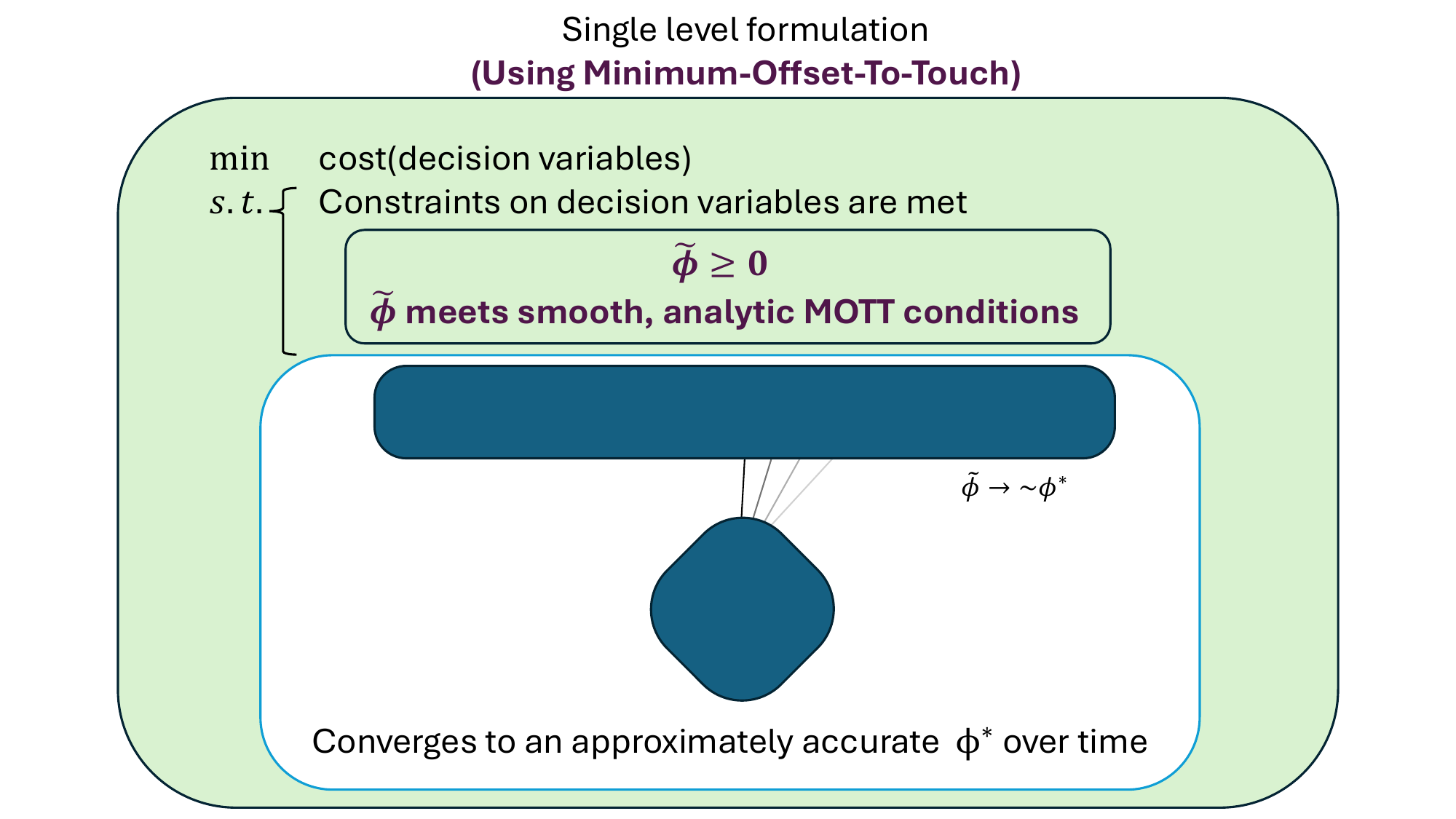}
\end{subfigure}
  \caption{
  Optimization-based methods for collision-aware trajectory optimization are traditionally bi-level (shown above in red), where a high level optimization problem calls a lower-level problem to calculate constrained values, like the signed distance between objects. In contrast, we propose Minimum-Offset-To-Touch (MOTT) conditions, which allow us to embed collision avoidance as smooth constraints with analytic form in the higher level problem, resulting in a single level optimization.  
  }
  \label{fig:intro_fig}
\end{figure}

In this paper, we present a formulation that allows us to compute an approximate signed distance function between sets, which we denote as `Minimum-Offset-To-Touch' (MOTT) (see Fig.~\ref{fig:intro_fig}). This can be used to enforce non-penetration constraints during trajectory optimization. Our method embeds computation of the signed distance function alongside the trajectory optimization problem. This is in contrast to the popular bi-level formulations, which require explicit calculation of distance at each solver iteration, as shown in Fig.~\ref{fig:intro_fig}. Other existing single-level optimization methods use a complementarity formulation (see~\cite{10440590}, for example) to impose distance function constraints during optimization which are, in general, difficult to solve leading to longer solve times. In contrast, MOTT conditions have continuous, smooth gradients, which allow us to enforce constraints on distances between smooth sets with faster compute efficiency. Through several numerical experiments, we present the computational benefits offered by the proposed method. 

\textbf{Contributions.}
\begin{itemize}
    \item We introduce a `Minimum-Offset-To-Touch' (MOTT) metric for the signed distance between collision bodies. 
    \item MOTT conditions allow us to derive computationally efficient, differentiable nonlinear equality conditions (compared to non-smooth complementarity conditions used in other approaches). These conditions can be embedded in a single-level optimization problem to enforce non-penetration and perform signed distance computation while performing trajectory optimization.
    \item Additionally, we propose a method for finding smooth semi-algebraic approximations of (non-smooth) polytopic sets, used in conjunction with our MOTT conditions to simplify optimization problems considering non-penetration of polytopic bodies.   
\end{itemize}
\section{Related Work}

Collision detection and avoidance has been thoroughly studied in robotics literature. There are various methods to compute collisions between bodies \cite{geibert1988fast, cameron1997enhancing, van2001proximity}. 
For example, the classic Gilbert–Johnson–Keerthi (GJK) algorithm \cite{geibert1988fast}. It computes the Minkowski difference of the two convex sets and estimates if the origin is inside this difference using iteratively computed simplices. The GJK algorithm and its variants \cite{montaut2024gjk++}
 have shown impressively fast collision computation. Some of them are used in the popular Flexible Collision Library (FCL) \cite{pan2012fcl} which is widely used in most physics engines. 
Another example is the Lin-Canny algorithm \cite{lin1991fast} which tracks the closest features (e.g., vertices, edges) between two convex polyhedra to check if collision is made. One of the issues for the aforementioned methods is that these solutions are not differentiable, which makes it difficult for gradient-based solvers (e.g., \cite{gill2005snopt}) to effectively use them for planning. 
Recently, Tracy et al. proposed DCOL \cite{tracy2023differentiable}, which enables differentiable collision computation for convex primitives. For iterative trajectory optimization solvers, all of the above methods require explicit computation of the collision values at each iteration.

Our work is also related to the vast literature on collsion-free motion planning in robotics.
Sampling-based methods such as RRT \cite{kuffner2000rrt} can efficiently plan collision-free trajectories. However, these methods do not consider kino-dynamic feasibility of generated motion plans. 
Model-free methods such as \cite{lin2021collision} can also design collision-free kinodynamically feasible trajectories, though training can be computationally demanding.
Graphs of Convex Sets \cite{marcucci2023motion} and Polytopic Action Set And Motion Planning \cite{jaitly2024PAAMP} form feasible motion planning as akin to a graph search problem. In these methods,  polytopic sets of feasible collision free motions are strung together to form sets of feasible longer-horizon motions. These methods require significant pre-computation to build the polytopic sets, and may require solving Mixed-Integer programs to find motion.



More recently, in \cite{le2023single} (SILICO) the authors introduced a method to solve for the collision values alongside the optimal solution to an optimization problem in a single level. However, this is achieved by introducing a number of non-smooth complementarity constraints, which can present gradient-based solvers with difficulties in converging.
In contrast, in this work, we present MOTT conditions, which are differentiable conditions with analytic expressions enforcing collision detection which do not require enforcing complementarity constraints to design collision-free trajectories, making it much easier for the solvers to converge.







\section{Background}
In this section, we present some relevant notation and background information which is useful in explaining the main results presented in this paper.
\subsection{Optimization-Based Collision Detection}

\subsubsection{Notation}
We first define some useful notation which is used throughout the paper:
\begin{itemize}
    \item $\mathbf{a} \leq \mathbf{b}$ (resp. $\frac{\mathbf{a}}{\mathbf{b}}$) denotes an elementwise inequality (resp. elementwise division) between vectors $\mathbf{a}$ and $\mathbf{b}$.
    \item $\mathbb{R}^{N}_-$ denotes the non-positive cone, or $x \in \mathbb{R}^N$ such that $x \leq 0$. $\mathbb{R}^{N}_+$ denotes the nonnegative cone.
    \item $\mathcal{A}_i = \{\mathbf{x} \in  \mathbb{R}^{N_x} | g_i(\mathbf{x}) \leq 0\}$ is a convex set where $g_i(\mathbf{x}) : \mathbb{R}^{N_x} \xrightarrow[]{} \mathbb{R}^{N_{g_i}}$ is a pre-defined function.
    \item The subscript $\cdot_{i,j}$ indicates that a value is associated with collision pair $\{\mathcal{A}_i, \mathcal{A}_j\}$.
    \item $\phi \in \mathbb{R}$ is a value associated with distance. $\phi_{i,j} \leq 0$ indicates that $\mathcal{A}_i$ and $\mathcal{A}_j$ are intersecting.
\end{itemize}

\subsubsection{Gilbert-Johnson-Keerthi Algorithm}
GJK~\cite{geibert1988fast} is a classic algorithm to compute collision between objects. The GJK algorithm effectively solves a `closest points problem', similar to the following,
\begin{equation}
    \min_{\mathbf{x}^i_{i,j}, \mathbf{x}^j_{i,j}} \frac{1}{2} \|\mathbf{x}^j_{i,j} - \mathbf{x}^i_{i,j}\|^2 \quad \text{s.t.} \quad \mathbf{x}^i_{i,j} \in \mathcal{A}_i, \quad \mathbf{x}^j_{i,j} \in \mathcal{A}_j \label{eq:GJK_QP}
\end{equation}

If the distance between the closest points is 0, then $\exists x \in \mathcal{A}_i \cap \mathcal{A}_j$ and the intersection is non-empty, reflecting that they are in penetration. Alternatively, if the minima is positive, the intersection is empty and the objects are not in penetration.

\subsubsection{DCOL}

In work considering Differentiable Collision Detection (DCOL~\cite{tracy2023differentiable}), penetration is evaluated by considering the `minimum scaling' amount that bodies must experience in order to be in contact. If they can be shrunk while preserving an intersection, the bodies are considered to be in collision. With $\alpha$ as the scaling factor, where $\alpha > 1$ relates to enlarging the set, $\alpha=0$ shrinks the set to a point, and $\alpha_{i,j}^*$ is the minimum scaling to touch, the value associated with distance, $\phi_{i,j}$, can be evaluated as $\phi_{i,j} = \alpha_{i,j}^* - 1$.

\subsubsection{KKT Conditions}

In this work, we are concerned with conditions that test the validity of a Minimum-Offset-To-Touch solution (introduced later). For this, we will require an understanding of the KKT conditions for optimality. 

The KKT conditions are sufficient and necessary conditions for a possible solution to a program like \eqref{eq:GJK_QP} to be optimal. The following are the KKT conditions of the closest points problem. Satisfying these conditions is equivalent to solving GJK.
\begin{subequations}
\begin{flalign} \label{gjk_kkt}
            &&  \text{find } {\mathbf{x}^i_{i,j}, \mathbf{x}^j_{i,j}, \lambda} \\
    \text{s.t.}    
            && (\mathbf{x}^j_{i,j} - \mathbf{x}^i_{i,j}) + \mathbf{\nabla_x g}_i(\mathbf{x}^i_{i,j}) \lambda_i = 0
             \label{eq:KKT_stationary_x1} \\
            && -(\mathbf{x}^j_{i,j} - \mathbf{x}^i_{i,j}) + \mathbf{\nabla_x g}_j(\mathbf{x}^j_{i,j}) \lambda_j = 0
             \label{eq:KKT_stationary_x2}\\
            && g_i(\mathbf{x}^i_{i,j}) \leq 0, \;\; g_j(\mathbf{x}^j_{i,j}) \leq 0    \label{eq:KKT_primary}\\
            && \lambda_i \geq 0, \;\; \lambda_j \geq 0              \label{eq:KKT_secondary} \\
            && \lambda_i^T g_i(\mathbf{x}^i_{i,j}) = 0, \;\; \lambda_j^T g_j(\mathbf{x}^j_{i,j}) = 0 
            \label{eq:KKT_complimentarity}
\end{flalign}
\end{subequations}
Here \eqref{eq:KKT_stationary_x1} and \eqref{eq:KKT_stationary_x2} are conditions of stationarity, \eqref{eq:KKT_primary} are conditions of primal feasibility, \eqref{eq:KKT_secondary} are conditions of dual feasibility, and \eqref{eq:KKT_complimentarity} are conditions of complimentary slackness. $\lambda$ are the dual variables in the KKT conditions.

\subsection{Superquadratics} \label{sec:superquad_introduced}
\begin{figure}[t]
    \centering
    \begin{subfigure}{0.485\linewidth}
        \adjustbox{trim=0.1\textwidth 0.1\textwidth 0.1\textwidth
        0.1\textwidth,clip}{\includegraphics[width=\textwidth]{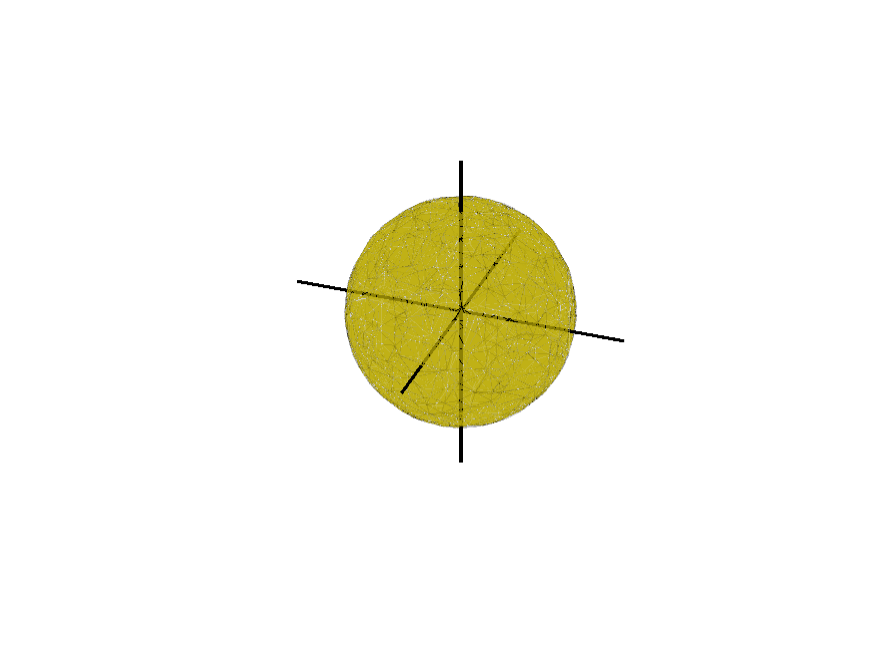}}
        \caption{$\rho = 1$}
    \end{subfigure}
    \begin{subfigure}{0.485\linewidth}
        \adjustbox{trim=0.1\textwidth 0.1\textwidth 0.1\textwidth 0.1\textwidth,clip}{\includegraphics[width=\textwidth]{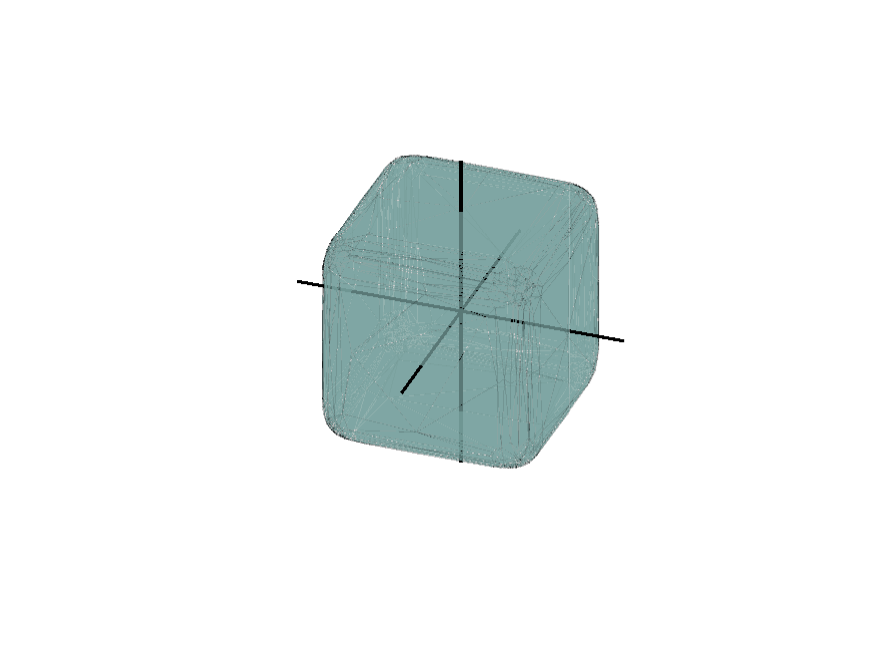}}%
        \caption{$\rho = 4$}
    \end{subfigure}
\caption{Superquadratic approximation of a unit-hypercube in $\mathbb{R}^3$. As $\rho$ increases, the approximation increases in accuracy.}
\label{fig:superquads}
\end{figure}

Our method utilizes smooth approximations of general polytopic sets. In order to do so, we employ 'superquadratics' (often refered to as superquadrics), seen in Fig. \ref{fig:superquads}. Superquadratics have been used to generate surfaces, largely for use in graphics applications \cite{Jaklic2000}\cite{ferreira2018superquad}. We will define superquadratic sets in $N_y$ dimensions to be $\tilde{R}^{N_y} = \{y \in \mathbb{R}^{N_y} \; | \; ||\mathbf{y}||^{2\rho}_{2\rho} \leq 1 \}$. $||\mathbf{y}||_{2\rho}$ is the $2\rho$ norm of $\mathbf{y}$, $(\sum_{k=0}^{N_y}(y_k ^ {2\rho}))^\frac{1}{2\rho}$. 

Superquadratics with positive integer values of $\rho$ can be visualized approximately as smooth, rounded hypercubes in $\mathbb{R}^{N_y}$. $\rho = 1$ results in a hyper-sphere of radius 1. As $\rho \xrightarrow{} \inf$, the superquadratic exactly matches $\{ -\mathbf{1} \leq \mathbf{y} \leq \mathbf{1}\}$.

\subsection{Trajectory Optimization}
Trajectory optimization problems considered in this work are solved by iteratively updating a guess for an optimal, feasible, solution. In this strategy, constraints are evaluated at each iteration. These constraints can span considerations of dynamics, kinematics, collision avoidance, etc.

In bi-level strategies, the signed distance ($\phi$) is algorithmically evaluated at each iteration. For a collision detection method like DCOL, this may require solving multiple convex programs at each iteration to evaluate the constraints. Single level strategies, on the other hand, may impose analytic constraints (often non-linear) on the decision variables that are only satisfied when $\phi$ is accurate. These strategies allow a mathematical program solver to converge towards $\phi$ values alongside the optimal values of other decision variables over the process of solving the optimization problem. 

Prior works such as~\cite{tracy2023differentiable},\cite{le2023single} have proposed single-level methods where the KKT conditions of the minimum scaling to touch problem are embedded as constraints in another optimization problem. These require enforcing complimentary slackness, which results in optimization with $N_{g_i} + N_{g_j}$ complementarity constraints for contact pair $\{\mathcal{A}_i, \mathcal{A}_j\}$. In situations like trajectory optimization, where the configuration ($q$) of the collision objects is variable, $g_i(x, q)$ may be non-convex, resulting in harder to enforce non-convex complementarity constraints for existing single level collision detection methods.

\section{Problem Statement}\label{sec:problem_statement}

In this section, we present our problem formulation. 
Our objective is to find a way to enforce constraints on the signed distance between objects easily.

We assume that the collision objects are closed, bounded, convex shapes $\mathcal{A}_i$ and $\mathcal{A}_j$ in $N_x$ dimensions (similar to previous work such as DCOL~\cite{tracy2023differentiable}). 

\begin{figure}[t]
    \centering
    \begin{subfigure}{0.49\linewidth}
        \includegraphics[width=\textwidth]{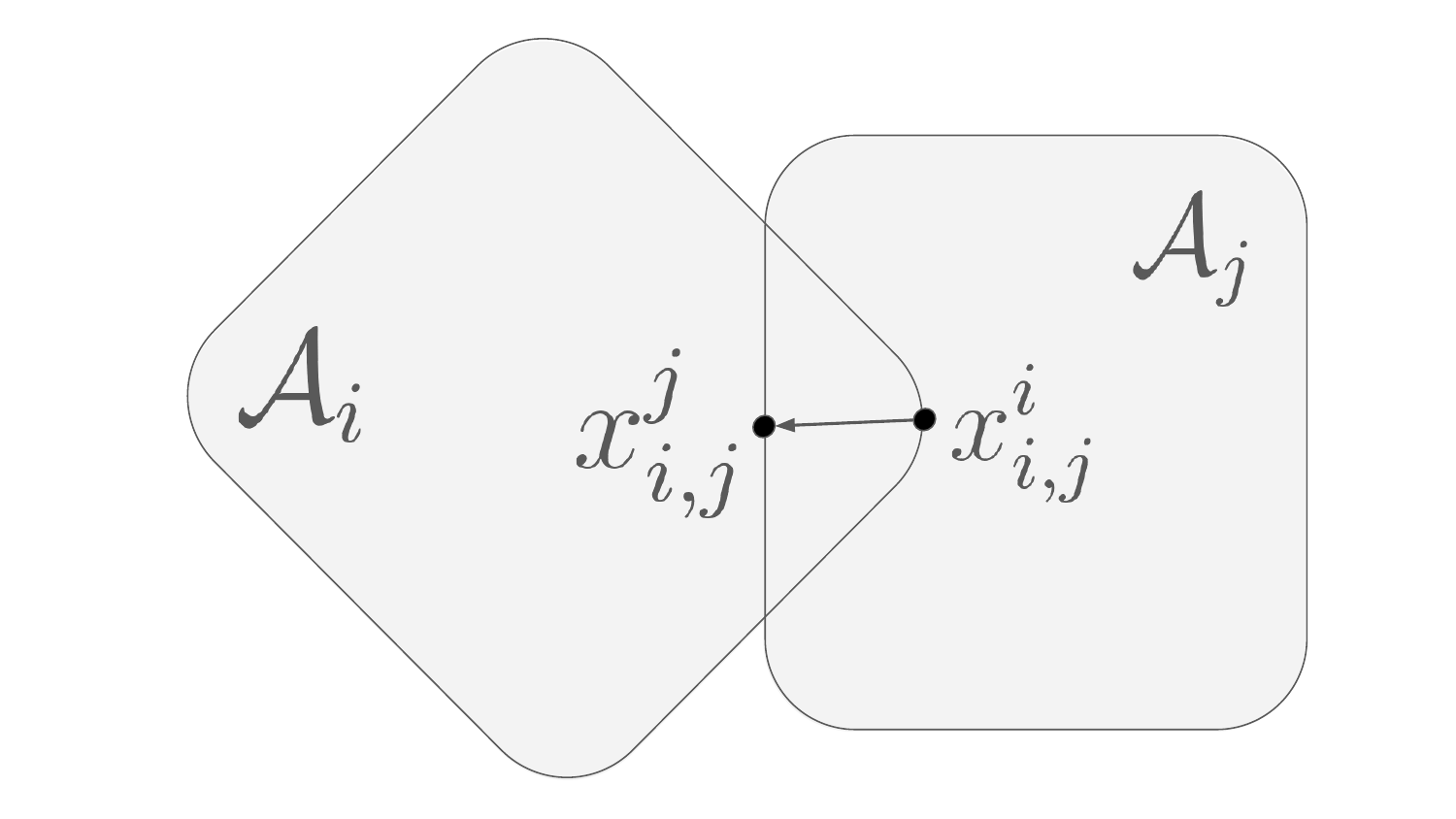}%
        \caption{Penetrating bodies.}\label{fig:objects_intersecting}
    \end{subfigure}
    \begin{subfigure}{0.48\linewidth}
        \includegraphics[width=\textwidth]{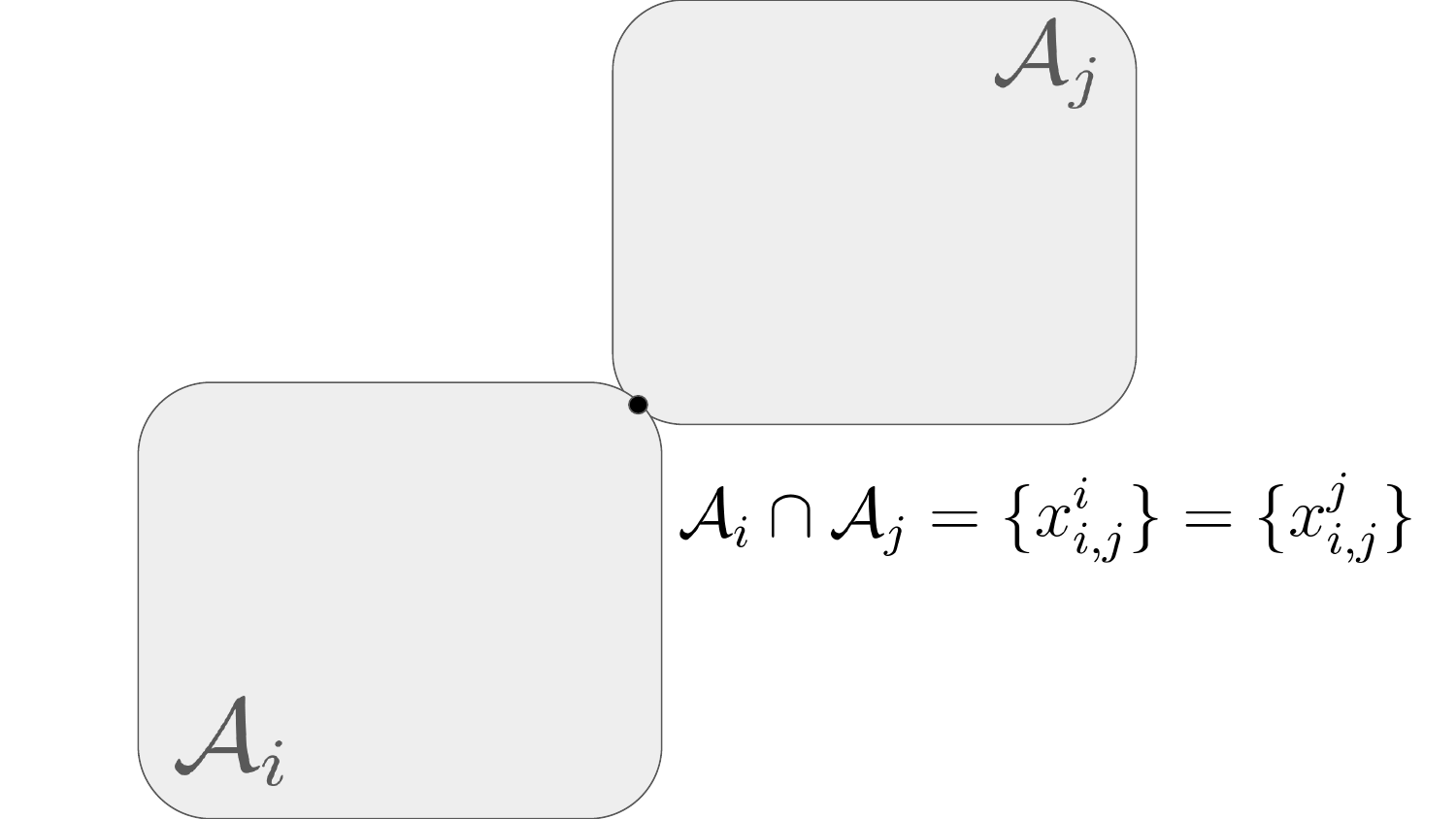}%
        \caption{Bodies that have intersection of cardinality 1. These objects are `touching'.}\label{fig:objects_touching}
    \end{subfigure}
    \caption{The `Minimum-Offset-To-Touch' distance is positive in the case of non-penetration, negative in the case of penetration, and 0 in the case that bodies are touching. Additionally shown are the points ($x_i, x_j$) that would be touching when the bodies are offset to touch.} 
    \label{fig:objects}
\end{figure}

We define our signed distance metric to be a `Minimum-Offset-To-Touch' (MOTT), a signed distance equivalent to the minimum amount that a body must be translated (without rotation) in any direction such that shapes are touching, but not overlapping. To touch, the intersection between translated objects must contain only points on the boundary of each set. We will consider non-flat surfaces, where this intersection is necessarily a single point. Examples of touching sets can be seen in Fig.~\ref{fig:objects_touching}. Offsets required to touch can be seen in Fig.~\ref{fig:objects_intersecting} and Fig.~\ref{fig:mott_conditions}.

Formally, at an `Offset-To-Touch',
\begin{itemize}
    \item $\mathbf{a}$ is a unit vector indicating a `direction of offset'.
    \item $\phi$ is the scalar valued signed offset distance.
    \item $(\mathcal{A}_i + \{\phi \mathbf{a} \}) \cap \mathcal{A}_j$ contains only points on the boundaries of each set.
    \item $\mathbf{x}^j_{i,j} \in \mathcal{A}_j$ is in the intersection of $\mathcal{A}_j$ and $\mathcal{A}_i + \{\phi \mathbf{a}\}$. This is a point at which the objects would be touching when $\mathcal{A}_i$ is offset by $\phi\mathbf{a}$.
    \item $\mathbf{x}^i_{i,j} \in \mathcal{A}_i$ is in the intersection of $\mathcal{A}_i$ and $\mathcal{A}_j - \{\phi \mathbf{a}\}$. This is a point at which the objects would be touching when $\mathcal{A}_j$ is offset by $-\phi\mathbf{a}$.
    \item $\phi \mathbf{a} = \mathbf{x}^j_{i,j} - \mathbf{x}^i_{i,j}$
\end{itemize}

At the MOTT solution, $(\phi, \mathbf{a}, \mathbf{x}^j_{i,j}, \mathbf{x}^i_{i,j})$, sets are touching, and $\phi^2$ is minimized.

\section{Method}

\begin{figure}[t]
  \centering\includegraphics[width=0.4\textwidth]{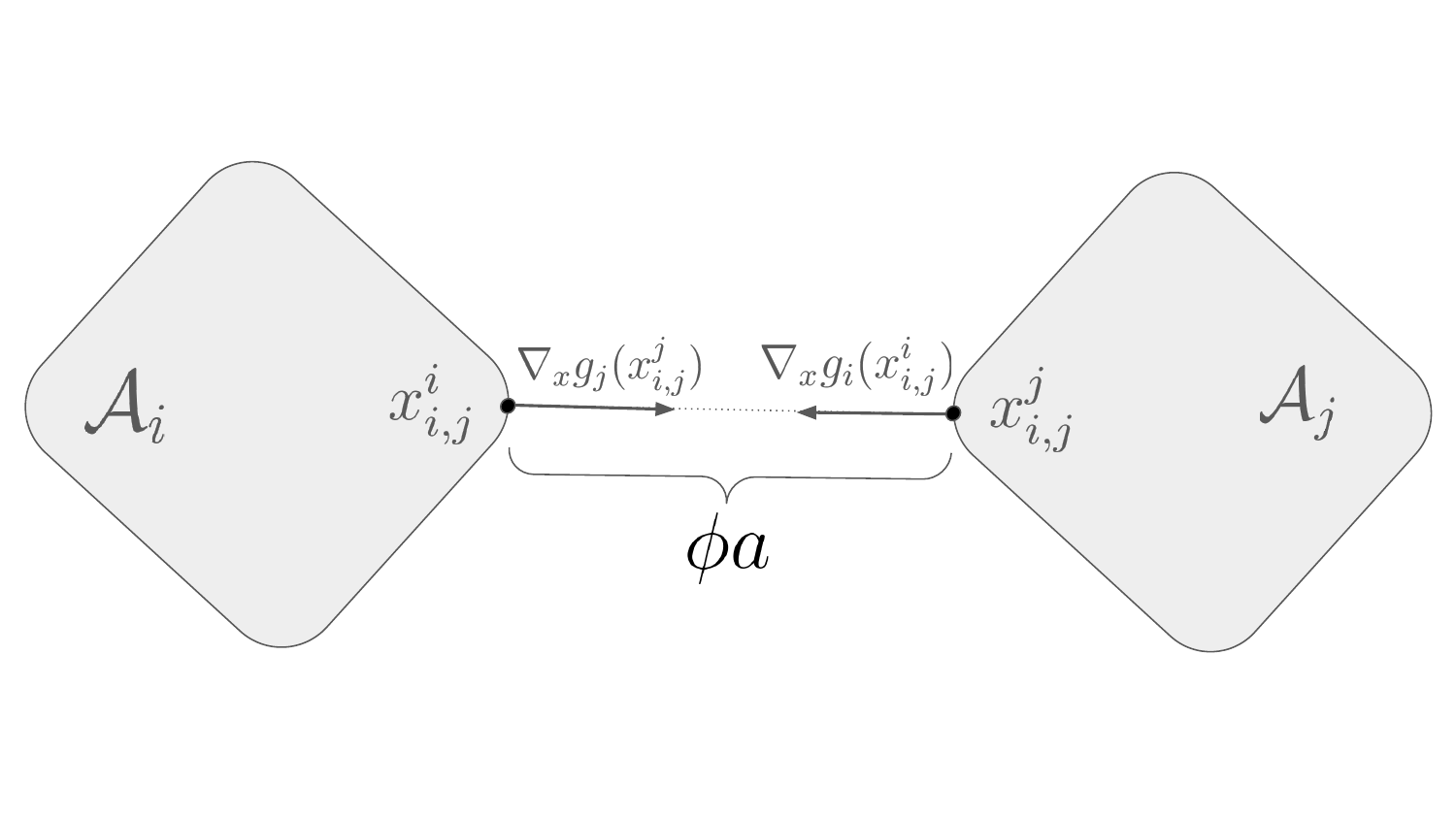}
  \caption{The `Minimum-Offset-To-Touch' conditions enforce that the surface normals ($\mathbf{\nabla_x g}_i(x_i)$, $\mathbf{\nabla_x g}_j(x_j)$) between two convex bodies are in opposite directions, and that the direction of offset ($\mathbf{a}$) is along this surface normal.}
  \label{fig:mott_conditions}
\end{figure}

In this section, we derive sufficient conditions for $( \phi, \mathbf{a}, \mathbf{x}^i_{i,j}, \mathbf{x}^j_{i,j} )$ to be a valid solution to the MOTT problem for shapes $\mathcal{A}_i$ and $\mathcal{A}_j$. We show that, for bodies with smooth surfaces, these are the smooth and algebraic conditions shown in Fig.~\ref{fig:mott_conditions}. We additionally introduce a method for approximating polytopic objects as smooth semi-algebraic sets.

\subsection{Collision Detection Between Smooth Bodies}

For collision detection, we specifically consider the case where $\mathcal{A}_i$ and $\mathcal{A}_j$ have smooth surfaces. For $\mathcal{A}_k$, we assume that $g_k(\mathbf{x})$ is a smooth, convex, scalar valued ($N_{g_k} = 1$) function.  

\subsubsection{Conditions for touching}
Recall the definition for touching from Sec.~\ref{sec:problem_statement}. Not every pair $\{\mathbf{x}^i_{i,j} \in \mathcal{A}_i, \mathbf{x}^j_{i,j} \in \mathcal{A}_j\}$ results in a valid offset value $\phi \mathbf{a} = \mathbf{x}^j_{i,j} - \mathbf{x}^i_{i,j}$ for touching. 
From our definition $\mathbf{x}^j_{i,j}$ is on the boundary of $\mathcal{A}_j$ and $\mathbf{x}^i_{i,j} = \mathbf{x}^j_{i,j} - \mathbf{a}\phi$ is on the boundary of $\mathcal{A}_i$. For scalar function valued constraints, this is defined as $g_j(\mathbf{x}^j_{i,j}) = 0$ and $g_i(\mathbf{x}^j_{i,j} - \mathbf{a}\phi) = g_i(\mathbf{x}^i_{i,j}) = 0$.

\newtheorem*{lemma1}{Lemma 1}
\begin{lemma1}

$(\phi \mathbf{a} = \mathbf{x}^j_{i,j} - \mathbf{x}^i_{i,j})$ defines a valid offset to touch if and only if 
\begin{subequations}
\begin{flalign}
    g_i(\mathbf{x}^i_{i,j}) = 
    g_j(\mathbf{x}^j_{i,j}) = 0 \label{eq:boundary_cond_1}\\
    \frac{\mathbf{\nabla_x g}_i(\mathbf{x}^i_{i,j})}{||\mathbf{\nabla_x g}_i(\mathbf{x}^i_{i,j})||} = -\frac{\mathbf{\nabla_x g}_j(\mathbf{x}^j_{i,j})}{||\mathbf{\nabla_x g}_j(\mathbf{x}^j_{i,j})||} \label{eq:normal_condition_1}
\end{flalign}
\end{subequations}

As stated prior, under the assumptions we have made the intersection $(\mathcal{A}_i + \{\phi \mathbf{a}\}) \cap \mathcal{A}_j$ contains a single point on the boundaries of both shapes, reflected in condition \eqref{eq:boundary_cond_1}. Note that, in cases where sets are non-smooth (like polytopes) this intersection may be set valued, instead of a single point. Our assumption of smoothness omits this case. 

By the separating hyperplane theorem \cite{boyd2004convex}, there necessarily exists at least one hyperplane tangent to both offset bodies at the contact point, which separates the interiors of the convex bodies. For smooth surfaces, this separating hyperplane is uniquely defined by the surface normal at the contacting point. This indicates that $\mathbf{\nabla_x g}_i( \mathbf{x}^i_{i,j})$ and $\mathbf{\nabla_x g}_j( \mathbf{x}^j_{i,j})$, the surface normals of each shape at the contact point, are in opposing directions if and only if the shapes (offset by $\phi \mathbf{a}$, where \eqref{eq:boundary_cond_1} is met) are touching, but the interiors are not overlapping. This gives \eqref{eq:normal_condition_1}. Thus, we can prove that our conditions are met if $\phi \mathbf{a}$ is an offset to touch.

If the conditions hold, we can state that $x^j_{i,j} = x^i_{i,j}+\phi\mathbf{a}$ is in the intersection, and on the boundaries, of $\mathcal{A}_i + \{\phi\mathbf{a}\}$ and $\mathcal{A}_j$ (from \eqref{eq:boundary_cond_1}) and that the interiors of the offset shapes are necessarily separated by the tangent hyperplane at $x^j_{i,j}$. Thus, any intersection is on the boundary of the shapes, and $\phi \mathbf{a}$ is an offset to touch.

\end{lemma1}
We use $\mathbf{\nabla_\mathbf{x}\hat{g}}_k(\mathbf{x})$ to indicate the directionality of the gradient of $g_k$, which is the surface normal. In the following lemma, we present the main result of this paper.

\newtheorem*{lemma2}{Lemma 2}
\begin{lemma2}
At the MOTT solution, $(\phi, \mathbf{a}, \mathbf{x}^j_{i,j}, \mathbf{x}^i_{i,j})$, the following conditions hold
\begin{subequations}
\begin{flalign}
g_i(\mathbf{x}^j_{i,j}) = 0, \;\;g_i(\mathbf{x}^i_{i,j}) = 0 \label{eq:c1_boundary}\\ 
\mathbf{\nabla_x g}_i(\mathbf{x}^i_{i,j}) = -\mathbf{\nabla_x g}_j(\mathbf{x}^j_{i,j})\label{eq:c1_gradient}\\ 
    (\mathbf{x}^j_{i,j} - \mathbf{x}^i_{i,j}) = \phi \mathbf{a} = \phi \mathbf{\nabla_x \hat{g}}_i( \mathbf{x}^i_{i,j}) = -\phi \mathbf{\nabla_x \hat{g}}_j( \mathbf{x}^j_{i,j})
\label{eq:c1_offset}
\end{flalign}
\end{subequations}

\end{lemma2}

\newtheorem*{proof1}{Proof 1}
\begin{proof}
Given the conditions for a valid `offset-to-touch' $(\mathbf{x}^i_{i,j}, \mathbf{x}^j_{i,j})$ the minimum offset to touch problem is the following:
\begin{subequations}
\begin{flalign}
    &&\underset{\phi, \mathbf{a}, \mathbf{x}^i_{i,j}, \mathbf{x}^j_{i,j}}{\min} \frac{1}{2} ||\mathbf{x}^j_{i,j} - \mathbf{x}^i_{i,j}||^2 \\
    \text{s.t.} && \phi \mathbf{a} = (\mathbf{x}^j_{i,j} - \mathbf{x}^i_{i,j}) \text{ is a valid offset to touch}
\end{flalign}
\end{subequations}
Or, using conditions \eqref{eq:boundary_cond_1}, \eqref{eq:normal_condition_1},
\begin{subequations}
\begin{flalign}
    &&\underset{\mathbf{x}^i_{i,j}, \mathbf{x}^j_{i,j}}{\min} \frac{1}{2} ||\mathbf{x}^j_{i,j} - \mathbf{x}^i_{i,j}||^2 \label{eq:mott_prog}\\ 
    \text{s.t.} && g_i(\mathbf{x}^i_{i,j}) = 0,\; g_j(\mathbf{x}^j_{i,j}) = 0 \\
    && \mathbf{\nabla_x \hat{g}}_i( \mathbf{x}^i_{i,j}) = - \mathbf{\nabla_x \hat{g}}_j( \mathbf{x}^j_{i,j}) \label{eq:mott_prog_end}
\end{flalign}
\end{subequations}
 
We wish to show that, if \eqref{eq:c1_boundary} -- \eqref{eq:c1_offset} hold for some solution, $(\phi, \mathbf{a}, \mathbf{x}^i_{i,j},\mathbf{x}^j_{i,j})$, then $(\phi, \mathbf{a}, \mathbf{x}^i_{i,j},\mathbf{x}^j_{i,j})$ is an optimal solution to \eqref{eq:mott_prog} -- \eqref{eq:mott_prog_end}.

Consider the KKT conditions of \eqref{eq:mott_prog} -- \eqref{eq:mott_prog_end},
\begin{subequations}
\begin{align}
&&\text{find } \mathbf{x}^i_{i,j}, \mathbf{x}^j_{i,j} \text{ s.t.} \label{eq:mott_kkt}\\
&& -(\mathbf{x}^j_{i,j} - \mathbf{x}^i_{i,j}) + \lambda_1 \mathbf{\nabla_x g}_i(\mathbf{x}^i_{i,j}) + \mathbf{\nabla^2_x g}^T_i( \mathbf{x}^i_{i,j})\mathbf{\lambda_3} = 0  \label{KKT_min_offset_1}\\
&&(\mathbf{x}^j_{i,j} - \mathbf{x}^i_{i,j}) + \lambda_2 \mathbf{\nabla_x g}_j( \mathbf{x}^j_{i,j}) + \mathbf{\nabla^2_x g}^T_j( \mathbf{x}^j_{i,j}) \mathbf{\lambda_3} = 0 \label{KKT_min_offset_2}\\
&& g_i(\mathbf{x}^i_{i,j}) = 0,\; g_j(\mathbf{x}^j_{i,j}) = 0 \label{eq:kkt_lemma_primal_1}\\
&& \mathbf{\nabla_x \hat{g}}_i(\mathbf{x}^i_{i,j}) = - \mathbf{\nabla_x \hat{g}}_j( \mathbf{x}^j_{i,j}) \label{eq:kkt_lemma_primal_2}
\end{align}
\end{subequations}

where $\nabla^2 g_k \in \mathbb{R}^{N_x \times N_x}$ indicates the Hessian of $g_k$. $\lambda_1, \lambda_2$ are arbitrary scalar free variables, and thus $\lambda_1 \mathbf{\nabla_x g}_i(\mathbf{x}^i_{i,j})$ is functionally equivalent to $\lambda_1 \mathbf{\nabla_x \hat{g}}_i(\mathbf{x}^i_{i,j})$.
Note that we are not required to enforce complimentarity. 

For $\mathbf{\lambda_3} = \mathbf{0}$, $\lambda_2 = \lambda_1 = \phi$, $ \phi \nabla_x \hat{g_i}(\mathbf{x}^i_{i,j}) = \mathbf{x}^j_{i,j} - \mathbf{x}^i_{i,j}$, we can see that \eqref{KKT_min_offset_1} and \eqref{KKT_min_offset_2} are equivalent to our condition, \eqref{eq:c1_offset}. \eqref{eq:kkt_lemma_primal_1} and \eqref{eq:kkt_lemma_primal_2} enforce primal feasibility, and directly reflect \eqref{eq:c1_boundary}, \eqref{eq:c1_gradient}. Thus, \eqref{eq:c1_boundary} -- \eqref{eq:c1_offset} are sufficient for enforcing the local optimality of a Minimum-Offset-To-Touch solution.
\end{proof}

As indicated in Fig. \ref{fig:mott_conditions}, \eqref{eq:c1_offset} indicates that $(\mathbf{x}^j_{i,j} - \mathbf{x}^i_{i,j})$ is `in the direction of' the surface normal, $\mathbf{\nabla_x g}_j(\mathbf{x}^j_{i,j})$. Note that, for objects that are not in penetration, the MOTT problem is equivalent to the closest-points problem.

\subsubsection{Minimum Offset To Touch for Polytopes} \label{sec:PolytopicMOTT}

In the case that we do wish to continue considering polytopic objects, our intuition would still hold. 
In this case, however, the `surface normal' would not be a unique vector value (as with our smooth body), but rather the normal cone at a point, $\mathbf{x}$. With abuse of notation, we state that \eqref{eq:c1_offset}, \eqref{eq:c1_gradient}, the surface normal constraints in the MOTT conditions, would be the following
\begin{align}
    \mathbf{a} \in \nabla_\mathbf{x} \mathcal{A}_j(\mathbf{x}^i_{i,j})
    ,\;\;
    -\mathbf{a} \in \nabla_\mathbf{x}\mathcal{A}_j(\mathbf{x}^j_{i,j})
\end{align}

Where $\nabla_x\mathcal{A}(x)$ indicates the normal cone of $\mathcal{A}$ at $x$, or a weighted sum of the surface normals of all facets of $\mathcal{A}$ that $x$ is a member of. 

Intuitively, stationarity and complimentary slackness constraints in the KKT conditions effectively enforce these conditions. Stationarity (\eqref{eq:KKT_stationary_x1}, \eqref{eq:KKT_stationary_x2}) enforces that $-\mathbf{\nabla_x f} = \mathbf{a} = \mathbf{A}^T \mathbf{\lambda}$ where $\mathbf{A}^T \mathbf{\lambda}$ denotes the weighted sum of the vectors normal to the facets of the polytope. Complimentary slackness, $\mathbf{\lambda}^T(\mathbf{Ax-b}) = 0$ as seen in \eqref{eq:KKT_complimentarity}, enforces that only the relevant subset of normal vectors are considered. Together, the conditions make $\mathbf{\nabla_x} \mathcal{A}(\mathbf{x}) = \{\mathbf{a} \;| \; \mathbf{a} = \mathbf{A}^T \mathbf{\lambda}, \;\;\mathbf{\lambda}^T(\mathbf{Ax-b}) = 0\}$, the normal cone to the point at $x$.  In considering scalar valued constraints, $g(\mathbf{x})$, we were able to make the assumption that $g(\mathbf{x}) = 0$ at the optimal solution and forgo the considerations of complimentary slackness that would traditionally be present in the KKT conditions of a similar problem. As there is only one `face' of the set (the boundary), we can state that the surface normal is always in a singular direction.



\subsubsection{Non-uniqueness}

The Minimum-Scaling-To-Touch problem, however, has non-unique locally optimal solutions, including the Maximum-Offset-To-Touch. 

\begin{figure}[htbp] 
  \centering
  \includegraphics[width=0.4\textwidth]{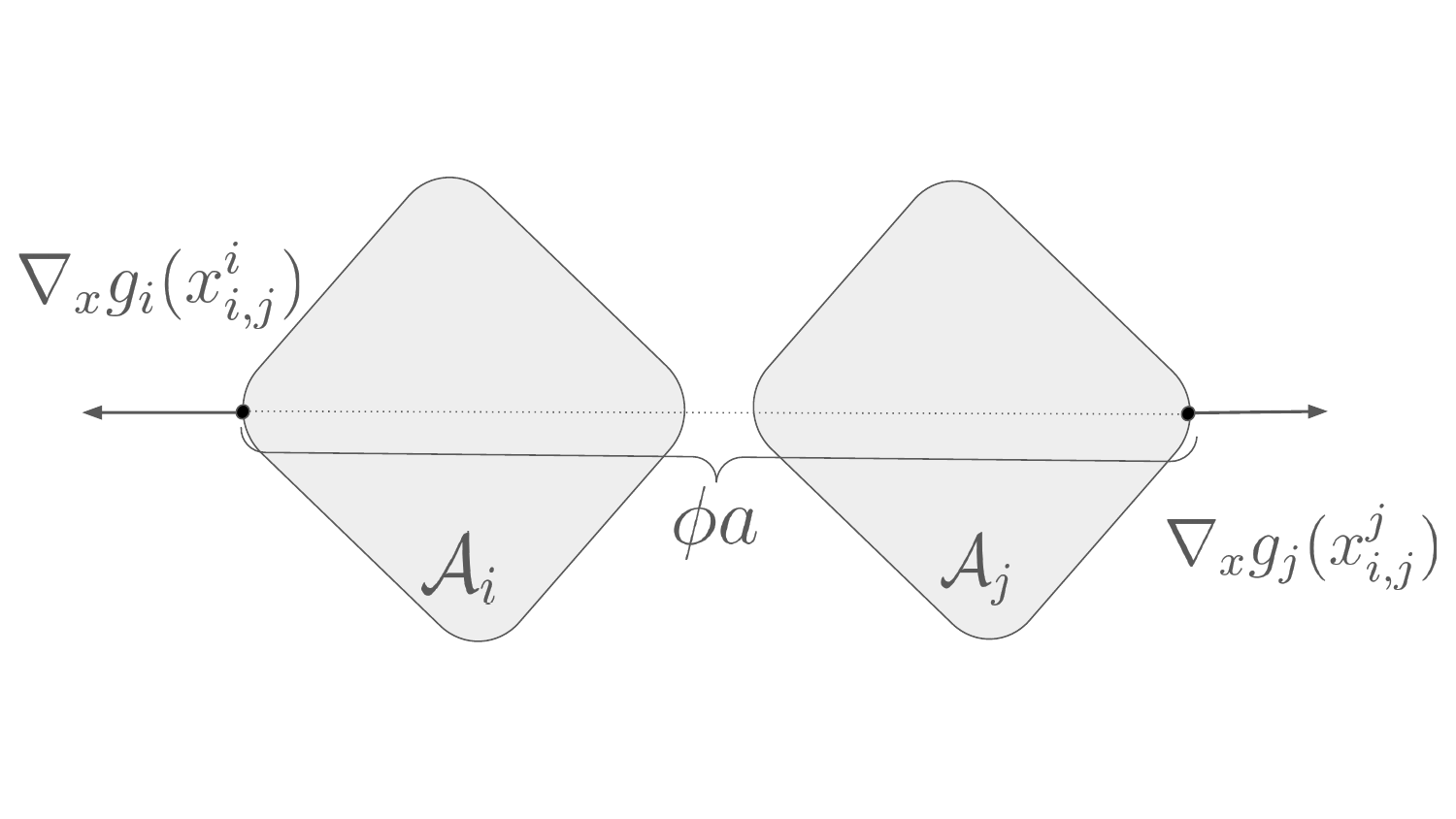}
  \caption{This is a sub-optimal solution for Minimum-Offset-To-Touch, even though it satisfies the conditions for local optimality.}
  \label{fig:inoptimal_sol}
\end{figure}

 The example in Fig. \ref{fig:inoptimal_sol} highlights an example of a `Maximum-Offset-To-Touch' solution, which finds the furthest points in the collision bodies. This solution is locally optimal to the Minimum-Offset-To-Touch program, and thus satisfies our nonlinear equality MOTT conditions. At such a solution, $\phi$ is negative, even though the objects are not in penetration. As can be seen in Fig. \ref{fig:inoptimal_sol}, the surface normals are pointing in opposing directions, away from the interior of each set. For this solution to be a `Minimum-Offset-To-Touch', the surface normals must point towards the interior of the opposing set. 
 
 So, we may consider an additional constraint of form 
\begin{equation}
(\mathbf{c}_j - \mathbf{c}_i) \cdot \mathbf{\nabla_x g}_i(\mathbf{x}^i_{i,j}) \geq 0 \label{eq:uniqueness}
\end{equation}
where $\mathbf{c}_j$ is in the interior of $\mathcal{A}_i$ and $\mathbf{c}_i$ is in the interior of $\mathcal{A}_j$. This enforces that $\mathbf{\nabla_x g}_i(\mathbf{x}^i_{i,j})$ is in the appropriate direction.

This condition may not be strictly necessary in practical applications, however. For instance, if the `Minimum-Offset-To-Touch' conditions are embedded in a trajectory optimization program enforcing non-penetration ($\phi \geq 0$), the `Maximum-Offset-To-Touch' solution -- where $\phi < 0$ -- is inherently excluded from the set of feasible solutions without the inclusion of \eqref{eq:uniqueness}.
 
\subsection{Smooth Approximations of Polyhedra} \label{sec:SmoothApprox}

When deriving the conditions for `Minimum-Offset-To-Touch', we made the assumption that each set considered was smooth, and that the constraint $g(\mathbf{x})$ was scalar valued. Non-smooth surfaces, and discontinuous gradients $\mathbf{\nabla_x \hat{g}(x)}$, result in non-smooth (and complementarity) constraints when enforcing validity of an offset value, $\phi$. This is not unique to our approach. As stated in \ref{sec:PolytopicMOTT}, approaches like SILICO \cite{le2023single} additionally require the enforcement of non-smooth complementarity constraints for non-smooth collision bodies. This is a restrictive assumption, however, as sets in question are not necessarily smooth. Instead, works have tended to approximate collision bodies with polytopes \cite{Deits2015IRIS}. In this subsection, we introduce the use of superquadratics to approximate generic polytopic sets as semi-algebraic sets with smooth surfaces and continuous gradients. 

\subsubsection{Smooth approximations by approximating the non-positive cone}

\begin{figure}%
\centering
\subfloat[Approximation the surface of the non-positive cone , noted in blue.]{\includegraphics[width=0.25\textwidth]{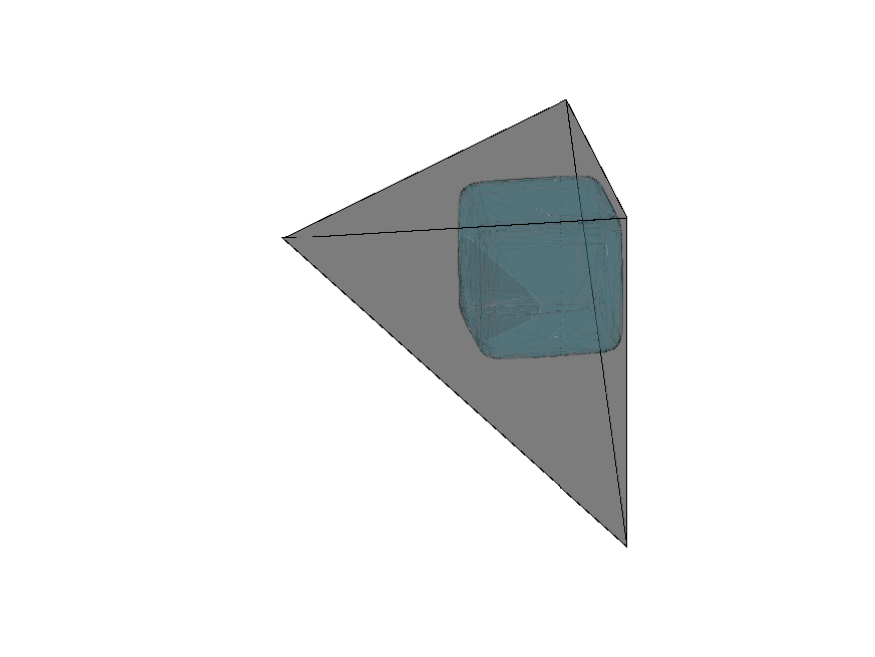}}
\subfloat[$\mathbf{y} = \mathbf{Ax-b}$ gives a hyperplane, noted in red.]{\includegraphics[width=0.25\textwidth]{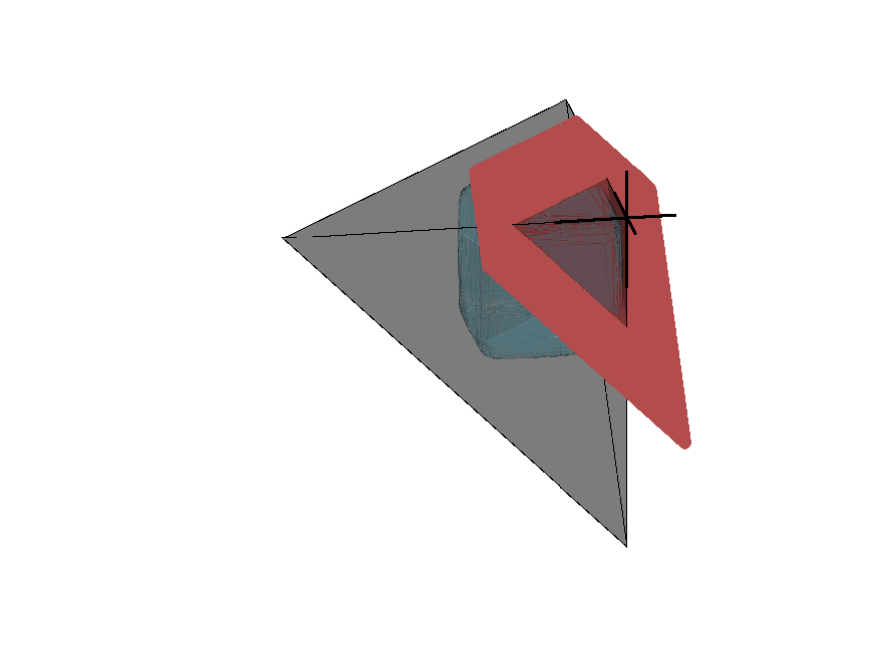}}\quad
\subfloat[The approximated semi-algebraic set, in blue, overlayed on the original polytope, outlined in black.]{\includegraphics[width=0.3\textwidth]{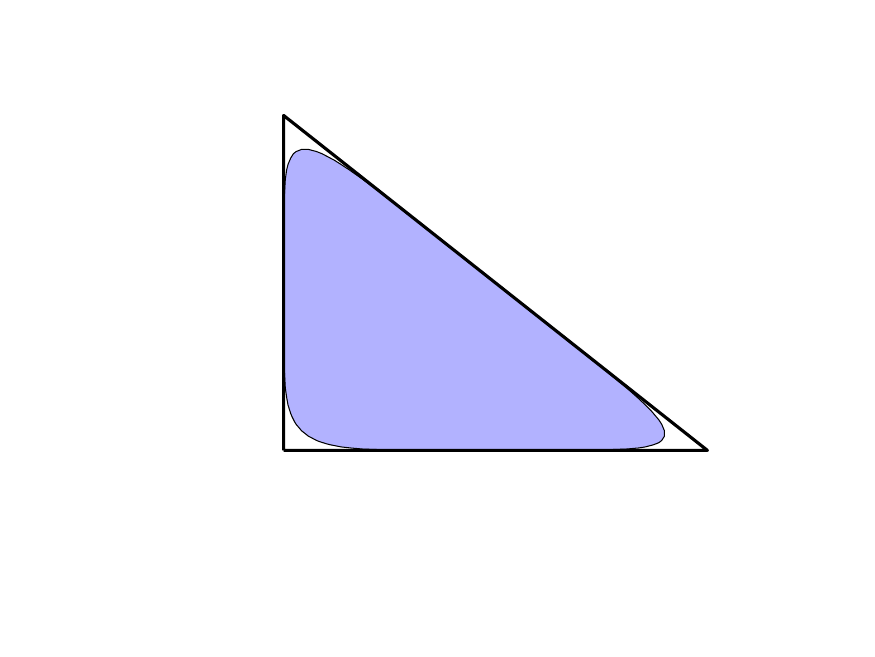}}
\caption{Superquadratic in the space $\mathbb{R}^{3}$. $\{\mathbf{y} = \mathbf{Ax-b}\} \cap \mathbb{R}^{N_y}_-$ represents the original polytope and $\{\mathbf{y} = \mathbf{Ax-b}\} \cap \tilde{\mathcal{R}}^{N_y}_-$ represents the approximated semi-algebraic set $\tilde{\mathcal{A}}$ (in blue).}
\label{fig:objects_intersection}
\end{figure}

A linearly constrained set generally takes the form $\mathcal{A} = \{x \in \mathbb{R}^{N_x} | \mathbf{y} = \mathbf{Ax-b}, \mathbf{y} \in \mathbb{R}^{N_y}_{-}\}$. We can understand a polyhedron to be the intersection between an affine subspace of $\mathbb{R}^{N_y}$ and the non-positive cone. Our approximation, $\tilde{\mathcal{A}}$, will depend on creating a smooth approximation of the non-positive cone, $$\tilde{\mathcal{R}}^{N_y}_- = \{\mathbf{y} | g(\mathbf{y}) \leq 0\} \subseteq \mathbb{R}^{N_y}_-$$ where $g(\mathbf{y}) \in \mathbb{R}^1$. The intersection between the same affine subspace and this smooth approximation (as seen in Fig.~\ref{fig:objects_intersection}), will yield a new set in $\mathbb{R}^{N_x}$ with a smooth boundary.

$
\tilde{\mathcal{A}} = \{ \mathbf{x}\in \mathbb{R}^{N_x} \;|\; 
\mathbf{y} = \mathbf{Ax-b},\;\;
\mathbf{y} \in \tilde{\mathcal{R}}^{N_y}_-\} 
$
gives our smooth approximation of the original polytope.


We will create an approximation of the non-positive cone using superquadratics. As mentioned in \ref{sec:superquad_introduced}, we will consider superquadratics of the form $||\mathbf{y}||^{2\rho}_{2\rho} \leq 1$. As $\rho \rightarrow \infty$,  $||\mathbf{y}||^{2\rho}_{2\rho} \leq 1$ becomes equivalent to $-\mathbf{1} \leq \mathbf{y} \leq \mathbf{1}$. 

Thus, $||\mathbf{y}||^{2\rho}_{2\rho} \leq 1$ approximates a hypercube in the space $\mathbb{R}^{N_y}$, centered at the origin. To approximate the non-positive cone, we will use 
\begin{align} 
\tilde{\mathcal{R}}^{N_y}_- =\{\mathbf{y}\in\mathbb{R}&^{N_y} \;| \;\left\| 2\frac{\mathbf{y}}{\mathbf{\bar{y}}} + 1 \right\| ^{2\rho}_{2\rho} \leq 1\}\label{eq:translated_superquad}
\end{align}
where $\mathbf{\bar{y}} \in \mathbb{R}^{N_y}$ and $\frac{\mathbf{y}}{\mathbf{\bar{y}}}$ represents an element-wise division.

\subsubsection{Choosing bounds for the superquadratic} \label{sec:choose_y}
 The resulting set approximates a hyper-rectangle within $\mathbb{R}_-^{N_y}$. The faces of the hyper-rectangle are congruent to the faces of the non-positive cone within the bounds $-\mathbf{\bar{y}} \leq \mathbf{y} \leq \mathbf{0}$. Thus, 
\begin{align}
\tilde{\mathcal{A}} = \{\mathbf{y} = \mathbf{Ax-b},\;\;
\left\| 2\frac{\mathbf{y}}{\mathbf{\bar{y}}} + 1 \right\| ^{2\rho}_{2\rho} \leq 1\}
\end{align}
is a valid approximation for the original polytope when $-\mathbf{\bar{y}} \leq Ax-b$. $\mathbf{\bar{y}}$ must be chosen to be sufficiently large, such that $-\mathbf{\bar{y}} \; \leq \; Ax-b \;\; \forall x \in \{x | Ax-b \leq 0\}$. 




\subsection{Trajectory Optimization}

In this subsection, we present the use of MOTT conditions in a collision avoidance problem. Due to the existence of kinematic considerations when performing trajectory optimization, we consider $\{\mathbf{x}_k | g_k(\mathbf{x}_k, \mathbf{q}) \leq 0\}$ as the approximated collision body in the world frame, where $\mathbf{q}$ is the configuration of the system. 

The program is as follows. 

\begin{align}
 \min && f(\mathbf{q}^0, \mathbf{q}^1, ..., \mathbf{q}^{T-1}) \label{eq:trajopt_problem}\\
 \text{s.t.} 
    && \mathbf{a^t_{i,j}} = \mathbf{\nabla_x \hat{g}}_i (\mathbf{x}^{t,i}_{i,j}, \mathbf{q}^t) \label{eq: trajopt_mott_start}\\ 
    &&
    \mathbf{a^t_{i,j}} = -\mathbf{\nabla_x \hat{g}}_j (\mathbf{x}^{t,j}_{i,j}, \mathbf{q}^t) \\
    && 
    \mathbf{x}^{t,j}_{i,j} - \mathbf{x}^{t,i}_{i,j} =\phi^t_{i,j}\mathbf{a}^t_{i,j}  \\      
    && g_i(\mathbf{x}^{t,i}_{i,j}, \mathbf{q^t}) = g_j(\mathbf{x}^{t,j}_{i,j},\mathbf{q^t}) = 0 \label{eq: trajopt_mott_end}\\ 
    &&
    \phi^t_{i,j} \geq 0\label{eq: trajopt_mott_nonpenetration}
\end{align}

$\mathbf{q}^t$ indicates the configuration at time step $t$, and $\mathbf{x}^{t,i}_{i,j}$ indicates the hypothetical contact point on $\mathcal{A}_i$ for pair $\{\mathcal{A}_i, \mathcal{A}_j\}$ in the world frame at time $t$. We enforce the constraints for each time step, for each contact pair.

\eqref{eq: trajopt_mott_start} -- 
\eqref{eq: trajopt_mott_end} are our Minimum-Offset-To-Touch conditions, while \eqref{eq: trajopt_mott_nonpenetration} enforces non-penetration of bodies $\mathcal{A}_i$ and $\mathcal{A}_j$. As stated in section \ref{sec:PolytopicMOTT}, enforcing non-penetration between polytopic objects requires enforcing that $\mathbf{a}$ is in the normal cone, which requires enforcing complementarity conditions. As such, we instead condsider  approximated polytopes, $\tilde{\mathcal{A}}$, as the relevant collision geometry. 
We calculate the superquadratic based semi-algebraic approximations of the collision bodies a-priori, and utilize them to evaluate $g(\mathbf{x})$ and $\nabla_x q(\mathbf{x})$. Our approximation ($\tilde{\mathcal{A}}_k$) is fully defined by $\mathbf{\bar{y}}_k$, $A_k$ and $b_k$ where $A_k, b_k$ are inherent to the original polytope, $\mathcal{A}_k$. We solve $N_y+1$ linear programs to find the bounds for $\mathbf{\mathbf{\bar{y}}}$ (as detailed in \ref{sec:choose_y}).

\section{Results}
In this section, we perform experiments to answer the following questions.
\begin{itemize}
    \item Can enforcing MOTT conditions efficiently find the correct Minimum Offset To Touch?
    \item Can our trajectory optimizer find collision-free trajectories in cluttered environments efficiently?
\end{itemize}

\subsection{Experiment Settings}
We implement our framework in C++ using SNOPT \cite{gill2005snopt} for solving \eqref{eq:trajopt_problem}. We implement our custom collision detection constraints within the Drake framework \cite{drake}. The computation is tested on a computer with Intel i7-1185G7. 


We test the performance of our method in both, a trajectory optimization scenario and for distance computation in a scenario with static objects. We use a Coin-OR Linear Program (CLP) solver to find $\mathbf{\bar{y}}_k$ a-priori. 

For trajectory optimization, we consider four distinct tasks with varying levels of complication in which we wish to facilitate collision free motion. We test our method, enforcing the Minimum-Offset-To-Touch conditions, and DCOL, recording the time it takes to solve. For our method, we also collect data on the average error in the solution `offset distance'.

The authors in~\cite{le2023single} (SILICO) primarily utilize the technique within a scenario where the configuration is fixed, yielding a problem with Linear Complimentarity constraints (an LCP). In trajectory optimization, this becomes a Nonlinear Complimentarity problem, and is significantly harder to solve. Thus, we do not consider the single level trajectory optimization problem using SILICO to find distance values. 

For each of the following, we set the initial guess by linearly interpolating between the initial and desired configurations. We attempt to minimize the distance from the goal at each of $T$ determined knot points, and impose a constraint, limiting the maximum velocity per knot point.

\subsection{Static Scene}
First, we implement an experiment solving for the Minimum-Offset-To-Touch with a static scene, where the configuration is pre-determined. This experiment will allow us to understand the time that a trajectory-optimizer may take to converge to an accurate solution for the Minimum-Offset-To-Touch, the approximation error, and how the two are related to $\rho$. We measure these values for 120 pairs of bodies randomly placed in space. 

For pairs not in penetration, the \% error is calculated as $\frac{\phi_{solution} - \phi_{accurate}}{\phi_{accurate}}$ where $\phi_{accurate}$ is the distance for the polytopes to touch (the GJK solution), and $\phi_{solution}$ is the offset for the smoothed approximated sets to touch.

The Minimum-Offset-To-Touch solution was used as an initial guess for the values of the offset-to-touch values. 

\begin{table}[!h]
\begin{center} 
\caption{The average solving time and accuracy for a program where we expect to find the Minimum-Offset-To-Touch, but not perform trajectory optimization.}\label{tab:rho_variation}
\begin{tabular}{ccc} 
 \toprule
$\rho$ & Solve Time (ms) & average \% error \\ \midrule
 1 & 24.30 & 18.34\\
 2 & 98.37 & 10.39\\
 3 & 58.48 & 6.81\\
 4 & 76.24 & 4.93\\
 5 & 29.30 & 3.81\\
 6 & 65.31 & 3.08\\
 7 & 32.25 & 2.80\\
 8 & 53.45 & 2.21\\
 9 & 54.85 & 1.93\\ \bottomrule
\end{tabular}
\end{center}
\end{table}

We additionally tested DCOL on the same scene. The average time to solve for the minimum-scaling-to-touch problem was 1.41 ms.

For the static scene, enforcing the Minimum-Offset-To-Touch conditions is not the ideal solution. This requires solving a program with nonlinear equality conditions, while solving for the Minimum Scaling To Touch constitutes a single quadratic program. The advantage of using the MOTT conditions, rather, is when an optimization has to be performed with multiple iterations, where the distance values change in each iteration. 

We can also see in Table~\ref{tab:rho_variation}, as expected, that as $\rho$ increases the approximation error decreases. This error comes from the fact that our smooth semi-algebraic approximation is an inexact inner approximation of the original smooth collision bodies. As $\rho$ increases, the gap between the approximated body and the original polytope decreases, and thus, as does the gap between the Minimum-Offset-To-Touch solutions. Additionally, the program solves for all considered values of $\rho$. 

\subsection{Trajectory Optimization}

\subsubsection{Freebody test}

The first trajectory optimization task considers two freely articulated cuboids that fly past each other. This enforces non-penetration for 1 collision pair.


\subsubsection{Bookshelf Scene}
\begin{figure}%
\centering
\subfloat[a collision-free trajectory for placing a book (red) between two static objects (shown in green).\label{fig:results_bookshelf}]{\includegraphics[width=0.25\textwidth]{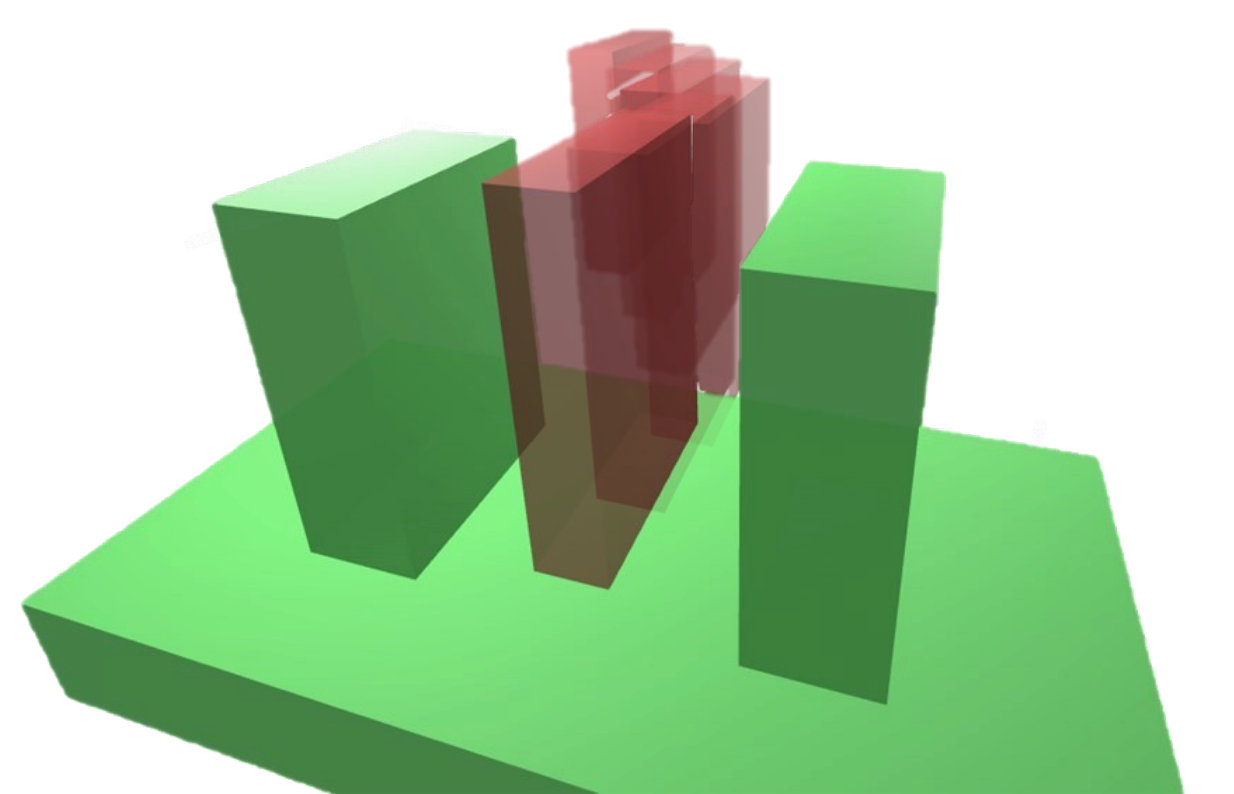}}
\subfloat[a collision-free trajectory for a drone (red) traversing through a messy scene. The drone maneuvers around the static obstacles (shown in green).\label{fig:results_drone}]{\includegraphics[width=0.25\textwidth]{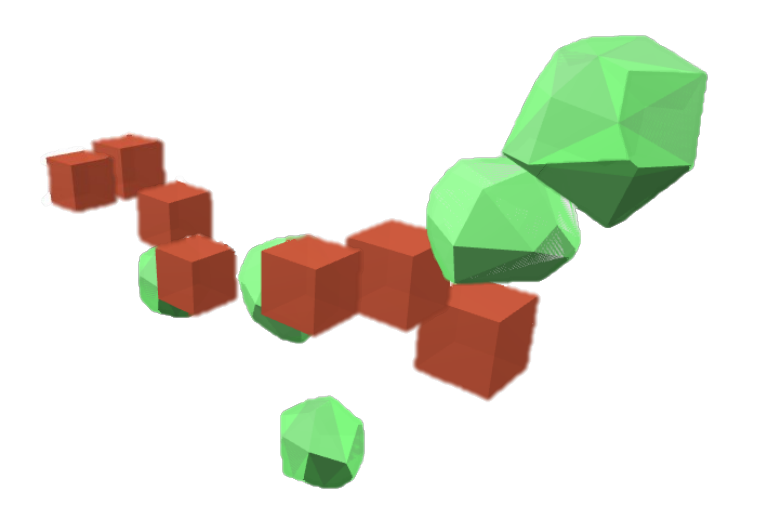}}
\caption{We consider scenarios of varying difficulty. Above are examples of the `drone scene' and the `bookshelf scene'.}
\end{figure}

The second task is a book placing task. We consider a book to be a freely articulated body, and wish to place it on a shelf with bookends on either side (Fig.~\ref{fig:results_bookshelf}). Linearly interpolating between the initial state and the desired final state results in a collision. We do not take into account dynamics considerations. This enforces non-collision for 3 collision pairs, with $\rho=3$.


\subsubsection{Drone Scene}

The third task considers a drone (approximated as a cuboid) traversing through a messy scene with polytopic obstacles (Fig.~\ref{fig:results_drone}). We consider the drone to be a freely moving body with 6 Degrees Of Freedom. We set the scene such that linearly interpolating between the initial and final states results in collision. This enforces non-penetration for 5 collision pairs, with $\rho = 3$. 


\subsubsection{Dual manipulator Scene}

\begin{figure}[t] 
  \centering
  \includegraphics[width=0.5\textwidth]{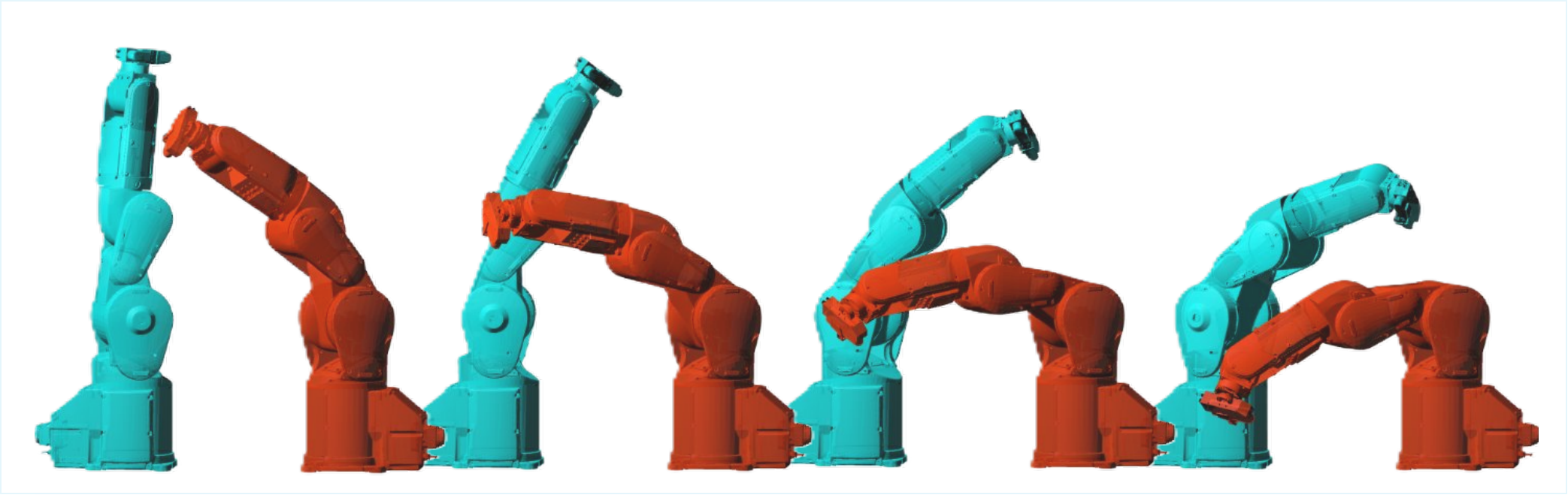}
  \caption{Snapshots of arm movement obtained using the proposed model in collision-free trajectory optimization. We attempt to find a collision-free trajectory between pre-determined poses.\label{fig:fig_result_robotarm}}
\end{figure}

In the final task, we consider a scene with two manipulators (Fig.~\ref{fig:fig_result_robotarm}), where the initial and final configurations are fixed. This enforces non-penetration for 15 collision pairs, with $\rho=5$.

\subsubsection{Trajectory Optimization Results}

\begin{table}[!h]
\begin{center}
\caption{
Solving times for trajectory optimization problems of varying difficulties. Enforcing the MOTT conditions is more efficient in problems that require more iterations, like the drone and dual manipulator scenes, as a solver converges to accurate distance values early. \label{tab:results_table}
}
\begin{tabular}{ccc} \toprule
    & MOTT & DCOL\\ \midrule
    Freebody Scene& 370.55 (ms)& 107.52 (ms)\\
    Bookshelf Scene& 419.04 (ms)& 111.94 (ms)\\
    Drone Scene& 2889.44 (ms)& 12128.40 (ms)\\
    Dual Manipulator Scene& 30.415(s) & 355.254(s)\\ \bottomrule
\end{tabular}
\end{center}
\end{table}

We see from Table~\ref{tab:results_table} that, for simpler problems which require less iterations to solve, the solver takes longer to converge to an accurate solution. It struggles to enforce the MOTT conditions while simultaneously taking large steps toward an optimal solution. 

For larger problems, on the other hand, enforcing non-penetration using a bi-level formulation takes significantly longer. Using minimum scaling to touch requires the explicit computation of the scaling at each iteration, which can tend to be expensive when all collision pairs, and all iterations, are taken into account. Enforcing the MOTT conditions in a single level requires that each iteration step of the SQP solver approximately preserve a valid value for $\phi$. 

\section{Conclusion}
In this paper, we proposed Minimum-Offset-To-Touch (MOTT) conditions for distance computation between convex bodies. These conditions are differentiable and computationally inexpensive conditions with analytic expressions that can be used to find an accurate metric for signed distance between convex bodies. Assuming that collision bodies are convex and smooth, we derive and verify that the minimum offset for two bodies to touch occurs along the surface normal of each body, and derive MOTT conditions to enforce this property. Using these conditions, a single-level trajectory optimization is able to efficiently and robustly find locally optimal trajectories for rigid bodies. It is noted that our approach does not require enforcing  complimentarity constraints allowing faster and robust computation. 
Furthermore, we introduce a method to approximate non-smooth, polytopic, objects using semi-algebraic sets to approximately find distance between non-smooth objects. 
Our method is extensively evaluated over various planning scenarios and outperforms the baseline method for trajectory optimization problems. 

We would like to extend this work to handle contact-rich manipulation tasks such as whole-body manipulation \cite{shirai2024linear, leve2024explicit} by incorporating contact dynamics as part of constraints. We would like to extend the use of the proposed method to model contact dynamics.

While we utilized this method to enforce non-penetration between physical objects in a trajectory optimization scenario, it can be used to enforce non-intersection of polytopes in higher dimensions as well. This can enable optimization methods to find polytopic inner approximations of non-convex sets, useful for methods like GCS \cite{marcucci2023motion} and Polytopic Action-Set And Motion Planning  \cite{jaitly2024PAAMP}. However, this is also left as a future exercise.



\bibliographystyle{IEEEtran}
\bibliography{main.bib}

\end{document}